\documentclass[11pt]{article}
\usepackage{emnlp2020}
\usepackage{times}
\usepackage{latexsym}
\usepackage{graphicx}              
\usepackage{multirow}
\usepackage{bm}
\usepackage{times}
\usepackage{amsmath}
\usepackage{enumerate}
\usepackage{CJK}
\usepackage{url}
\usepackage{booktabs} 
\usepackage{footnote}
\usepackage{tabularx}
\usepackage{algorithm}
\usepackage{algorithmic}
\usepackage{float}
\usepackage{color}
\usepackage{arydshln}
\usepackage{bm}
\usepackage{amssymb}

\usepackage{colortbl} 
\usepackage{multirow} 
\usepackage{multicol}
\usepackage{colortbl}
\usepackage{arydshln}

\aclfinalcopy 

\title{Semantic Role Labeling Guided Multi-turn Dialogue ReWriter}

\author{Kun Xu$^{1}$, Haochen Tan$^{2}$, Linfeng Song$^{1}$, Han Wu$^{2}$, Haisong Zhang$^{1}$, Linqi Song$^{2}$, Dong Yu$^{1}$\\
         $^{1}$Tencent AI Lab \\ 
	 $^{2}$City University of Hong Kong \\
	\{\texttt{kxkunxu},\texttt{lfsong},\texttt{hansongzhang},\texttt{dyu}\}\texttt{@tencent.com}\\
	\{\texttt{haochetan2-c},\texttt{hanwu32-c}\}\texttt{@my.cityu.edu.hk}\\
	\{\texttt{linqi.song}\}\texttt{@cityu.edu.hk}
}

\date{}

\begin{document}
\maketitle
\begin{CJK*}{UTF8}{gbsn}

\begin{abstract}
For multi-turn dialogue rewriting, the capacity of effectively modeling the linguistic knowledge in dialog context and getting rid of the noises is essential to improve its performance.
Existing attentive models attend to all words without prior focus, which results in inaccurate concentration on some dispensable words.
In this paper, we propose to use semantic role labeling (SRL), which highlights the core semantic information of \emph{who} did \emph{what} to \emph{whom}, to provide additional guidance for the rewriter model.
Experiments show that this information significantly improves a RoBERTa-based model that already outperforms previous state-of-the-art systems.
\end{abstract}

\section{Introduction}

Recent research \cite{vinyals2015neural,li2016diversity,serban2017hierarchical,zhao2017learning,shao2017generating} on dialogue generation has been achieving impressive progress for making single-turn responses,
while producing coherent multi-turn replies still remains extremely challenging.
One important factor that contributes to this difficulty is coreference and
information omission, where mention is dropped or replaced by a pronoun for simplicity.
These phenomena dramatically introduce the requirements for long-distance reasoning, as they frequently occurred in our daily conversations, especially in pro-drop
languages like Chinese and Japanese.

To tackle these problems, sentence rewriting was introduced to ease the burden of dialogue models by simplifying the multi-turn dialogue modeling into a single-turn problem.
Several approaches \cite{su2019improving,zhang2019filling,elgohary-etal-2019-unpack} have been proposed to address the rewriting task.
Conceptually, these models follow the conventional encoder-decoder architecture that first encodes the
dialogue context into a distributional representation and then decodes it to the rewritten utterance.
Their decoders mainly use global attention methods
that attends to all words in the dialogue context without prior focus, which may result in inaccurate concentration on some dispensable words.
We also observe that the accuracy of their models significantly decreases when working on long dialogue contexts.
This observation is expected since if the text is lengthy, it would be quite difficult for deep learning models to understand as it suffers from noise and pays vague attention to the text components.

\begin{table}[t!]
\small
\centering
\begin{tabular}{l|l}
\toprule[0.8pt]
Utterance 1 & 需要粤语 \\
 & (I may need Cantonese.) \\
Utterance 2 & $\underline{\text{粤语}}$$_{\textbf{ARG0}}$是$\underline{\text{普通话}}$$_{\textbf{ARG1}}$吗 \\
& (Is Cantonese Mandarin ?) \\
Utterance 3 & $\underline{\text{不算}}$$_{\textbf{predicate}}$吧 \\
& (Maybe Not.) \\
\textbf{\textcolor{blue}{Utterance 3$^\prime$}} & \textbf{\textcolor{blue}{粤语不算普通话吧}} \\
& (Cantonese may be not Mandarin.) \\
\toprule[0.8pt]
\end{tabular}
\caption{One example of multi-turn dialogue. The goal of dialogue rewriting is to rewrite utterance 3 into 3$^\prime$.}
\label{tab:intro_examples}
\vspace{-3mm}
\end{table}

\begin{figure*}[ht!]
    \centering
    \includegraphics[width=1.0\linewidth]{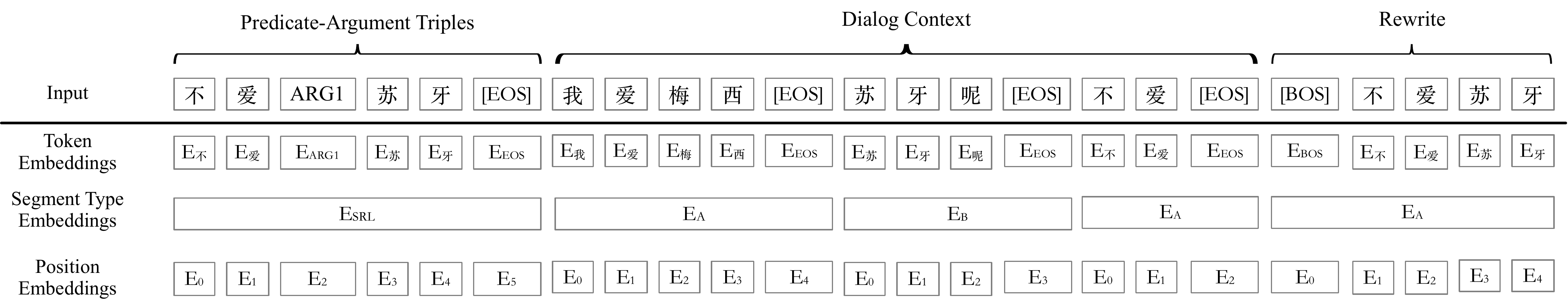}
    \caption{The input representation of a running example. We should point out that some tuples that do not contain words in the rewritten utterances could also be used as input predicate-argument triples.}
    \label{fig:model}
    \vspace{-3mm}
\end{figure*}
Motivated by these observations, we propose to incorporate the information of Semantic role labeling (SRL) \cite{gildea2002automatic,palmer2010semantic} to improve sentence rewriting.
SRL is broadly used to identify the predicate-argument structures of a sentence, where these structures could capture the main semantic information of \emph{who} did \emph{what} to \emph{whom}.
As a result, we believe that it can pick out the important words,
which are semantically most related to the utterance that needs to be rewritten.
As shown in Table~\ref{tab:intro_examples},
our SRL system is able to find that the \texttt{\small ARG0} and \texttt{\small ARG1} of ``{\small 不算}''(is not) are ``{\small 粤语}''(Cantonese) and ``{\small 普通话}''(Mandarin), respectively.
Consequently, our rewriting model can correctly generate the correct output (utterance 3$^\prime$), which covers all dropped information.
We can see that SRL can guide our rewriting model to focus on the semantically important words in the dialogue history, especially the omitted information that appears in previous turns.

For more details, we first take an SRL parser to recognize the predicate-argument (PA) structures from dialog contexts, before encoding that semantic information into our model.
Since conventional SRL benchmarks only contain sentence-level annotations, existing pretrained SRL parsers \cite{khashabi2018cogcompnlp,gardner2018allennlp} can 
fail to extract the cross-turn PA structures in dialogues.
To address this problem, we extend the traditional SRL to the conversational scenario by additionally annotating a dialogue dataset with standard SRL labels.

Our rewriting model is based on a pre-trained RoBERTa model \cite{liu2019roberta} that takes the outputs of SRL parsing and dialogue history as its inputs, before generating rewriting outputs word by word. 
Experimental results show that even without the SRL information, our model already outperforms previous state-of-the-art models by a large margin.
Augmenting the SRL information, the model performance is further improved significantly \emph{without} adding any new parameters.

\section{Task Definition}
Formally, an input for dialogue rewriting is a dialogue session $c = (u_{1}, ..., u_{N})$ of $N$ utterances, and $u_{N}$ is the most recent utterance that needs to be revised.
The output is $r$, the resulting utterance after recovering all coreference and omitted information in $u_{N}$.
Our goal is to learn a model that can automatically rewrite $u_{N}$ based on the dialogue context.

\section{Model}
Given a dialogue context $c$, we first apply an SRL parser to identify the predicate-argument structures $z$; then conditioned on $c$ and $z$, the rewritten utterance is generated as $p(r|c, z)$.
The backbone of our infrastructure is similar to the transformer blocks in \newcite{dong2019unified},
which supports both bi-directional encoding and uni-directional decoding flexibly via specific self-attention masks.
Specifically, we concatenate $z$, $c$ and $r$ as a sequence, feeding them into our model for training; during decoding, our model takes the $z$ and $c$ before generating the rewritten utterance word by word.
Our model uses a pre-trained Chinese RoBERTa \cite{liu2019roberta} for rich features.

\begin{table*}[t!]
\centering
\begin{tabular}{l|c|c|c|c|c|c|c}
\toprule[0.8pt]
& BLEU-1 & BLEU-2 & BLEU-4 & ROUGE-1 & ROUGE-2 & ROUGE-L & EM \\ \hline
Trans-Gen & 78.18 & 70.31 & 51.85 & 83.1 & 67.84 & 81.98 & 24.12 \\
Trans-Pointer & 83.22 & 78.32 & 64.08 & 87.89 & 77.94 & 86.88 & 36.54 \\
Trans-Hybrid & 82.92 & 77.65 & 62.54 & 87.59 & 76.91 & 86.66 & 35.03\\
\newcite{su2019improving} & 85.41 & 81.67 & 70.00 & 89.75 & 81.84 & 88.56 & 46.33 \\
\hline
BERT & 88.21 & 85.17 & 75.64 & 90.73 & 84.35 & 89.47 & 57.36\\
BERT + SRL &  \\
\quad w/ Bi-mask & 88.89 & 85.88 & 76.36 & 90.92 & 85.00 & 89.72 & 58.36 \\
\quad w/ Triple-mask & \textbf{89.66} & \textbf{86.78} & \textbf{77.76} & \textbf{91.82} & \textbf{85.87} & \textbf{90.52} & \textbf{60.49} \\
BERT + Partial-SRL & 89.46 & 86.57 & 77.75 & 91.60 & 85.60 & 90.50 & 59.15\\
\hline
BERT + Gold-SRL & 93.34 & 91.38 & 84.97 & 94.94 & 90.45 & 93.86 & 71.96 \\
\toprule[0.8pt]
\end{tabular}
\caption{Evaluation results on the datset of \newcite{su2019improving}.}
\label{tab:results}
\end{table*}

\subsection{Conversational SRL}
SRL has long been treated as a sentence-internal task, and its major benchmarks \cite{carreras2005introduction,pradhan2013towards} contains only sentence-level annotations.
We extend SRL to fit the conversational scenario by allowing SRL parsers to search for potential arguments over the whole conversation.
As there is no publicly available data with paragraph-level SRL annotations, we directly annotate inter- and cross-utterance arguments for predicates on a public dialogue dataset, Duconv \cite{wu-etal-2019-proactive}\footnote{More annotation details could be found in the Appendix.}.
Specifically, we annotated 3,000 dialogue sessions, including 33,673 predicates in 27,198 utterances.
Among them, 21.89\% arguments are not in the same turn with their predicates, respectively.
Considering existing standard SRL benchmarks may also be helpful, we first pre-train our SRL model \cite{shi2019simple} on the training set of CoNLL~2012 (117,089 examples) and fine-tune it on our annotations.
In our experiments, we employ this conversational SRL model to recognize the predicate-argument structures for the dialogue context.

\subsection{Input Representation for ReWriter}
For each token, its input representation is obtained by summing the embeddings for word, semantic role and position.
One example is shown in Figure~\ref{fig:model} and details are described in the following:\\
$\bullet$ The input is the concatenation of PA structures, dialog context, and rewritten utterance.
Note that a PA structure is essentially in a tree format, where the root is a predicate and its children are corresponding semantic arguments.
For the linearization, we decomposing each PA structure into several triples
of the form $<$\textit{predicate}, \textit{role}, \textit{argument}$>$ and concatenate them in a random order.
A special end-of-utterance token (i.e., [EOS]) is appended to the end of each utterance for separation.
Another begin-of-utterance token (i.e., [BOS]) is also added at the beginning of the rewritten utterance.
The final hidden state of the last token in the final layer is used to predict the next token during generation.\\
$\bullet$ We expand the segment-type embeddings of BERT to distinguish different types of tokens.
In particular, the type embedding E$_{A}$ is added for the rewritten utterance, as well as dialogue utterances generated by the same speaker in the context; the type embedding E$_B$ is used for the other speaker; E$_{SRL}$ is used as the type embedding of the tokens in predicate-argument triples.
Position embeddings are added according to the token position in each utterance.
The input embedding is the summation of word embedding, segment embedding, and position embedding.

\subsection{Attention Mask}
Similar to TransferTransfo \cite{wolf2019transfertransfo}, we apply a future mask on the rewritten sequence,
that is, the tokens in the rewritten utterance only attend on previous tokens in self-attention layers.
Recall that, we linearize a PA structure into a concatenation sequence of triples.
Since these triples are randomly ordered, it may inevitably introduce noisy information when using a sequence encoder.
To better reflect its structural information, we elaborate the attention mask on PA sequence:
the tokens in the same PA triple have bidirectional attentions while tokens in different PA triples
can not attend each other.
And the position embeddings of tokens in the PA sequence are added according to their positions in each distinct triple rather than the total PA sequence.
In experiments, we find using these two designs help our model to more efficiently use the SRL information.
We leave a more detailed discussion in Session~\ref{exp}.

\subsection{Training}
We employ the NLL loss to train our model:
\begin{equation}
\nonumber
    \mathcal{L}  = - \sum_{t=1}^{T} \log p(r_t|c, z, r_{< t}; \bm{\theta})
\end{equation}
where $\bm{\theta}$ represents the model parameters, $T$ is the length of the target response $r$,
and $r_{< t}$ denotes previously generated words.

\section{Experiments}
\label{exp}
We evaluate our model on two rewrite datasets, which are built by \newcite{su2019improving} and \newcite{cai2019retrieval}.
Both of these two datasets are generated by crawling multi-turn conversational data from several popular Chinese social media platforms.
Specifically, the dataset of \newcite{su2019improving} contains 17,890 examples, which are further split as 80\%/10\%/10\% for training/development/testing, respectively.
The dataset of \newcite{cai2019retrieval} contains 204k examples, where 194k/5k/5k are for training/developement/testing.

The hyper-parameters used in our model are listed as follows.
The network parameters of our model are initialized using RoBERTa.
The batch size is set to 32.
We use Adam \cite{kingma2014adam} with learning rate 5e-5 to update parameters.

\begin{table}[t!]
\centering
\begin{tabular}{l|c|c|c|c}
\toprule[0.8pt]
& B1 & B2 & R1 & R2  \\ \hline
Trans-Pointer & 84.70 & 81.70 & 89.00 & 80.90 \\
\hline
BERT & 85.21 & 82.51 & 89.53 & \textbf{83.18} \\
BERT + SRL & \textbf{85.77} & \textbf{82.85} & \textbf{89.59} & 83.08  \\
\toprule[0.8pt]
\end{tabular}
\caption{Evaluation results on the datset of \newcite{cai2019retrieval}. B$_{n}$ represents n-gram BLEU score and R$_n$ represents n-gram ROUGE score.}
\label{tab:results_2}
\end{table}

\noindent\textbf{Results and Discussion.}~~
Following previous works, we used BLEU, ROUGE, and the exact match score (EM) (the percentage of decoded sequences that exactly match the human references).
We implemented three baselines that use the same transformer-based encoder but differ in the choice of the decoder.
Specifically, \textit{Trans-Gen} uses a pure generation decoder which generates words from a fixed vocabulary;
\textit{Trans-Pointer} applies a pure pointer-based decoder \cite{vinyals2015pointer} which can only copy the word from the input;
\textit{Trans-Hybrid} uses a hybrid pointer+generation decoder as in \newcite{see2017get}, which can either copy the words from the input or generate words from a fixed vocabulary.
Table~\ref{tab:results} and Table~\ref{tab:results_2} summarizes the results of our model and these baselines.

We can see that even without the SRL information, our model still significantly outperforms these baselines on two datasets,
indicating that adapting a pre-trained language model could greatly improve the performance of such a generation task.
We can also see that the model with the pointer-based decoder achieves better performance than the generation-based and the hybrid one, which is similar to the observation as in \newcite{su2019improving}.
This result is expected since there is a high chance the coreference or omission could be perfectly resolved by only using previous dialogue turns.
In addition, we find that incorporating the SRL information can further improve the performance by at 1.45 BLEU-1 and 1.6 BLEU-2 points, achieving the state-of-the-art performances on the dataset of \newcite{su2019improving}.

Let us first look at the impact of attention mask design on our model.
To incorporate the SRL information into our model, we view the linearized predicate-argument structures as a regular utterance (say $u_{pa}$) and append it in the front of the input.
We experimented with two choices of attention masks.
Specifically, the first one is a bidirectional mask (referred as Bi-mask), that is, words in $u_{pa}$ could attend each other;
the second one (referred as Triple-mask) only allows words to attend its neighbors in the same triple, i.e.,
words in different triples are not visible to each other.
From Table~\ref{tab:results}, we can see that the latter one is significantly better than the first one.
We think the main reason is that the second design independently encode each predicate-argument triple,
which prevents the unnecessary triple-internal attentions, better mimicking  the SRL structures.

Since our framework works in a pipeline fashion, one bottleneck of our system can lie in the performance of the SRL parser.
One natural question is how accurate our SRL parser can be and how much performance improvement for the rewriter model we could have by introducing the SRL information.
To investigate this, we employ a conventional SRL parser\footnote{This SRL parser is trained on the CoNLL-2012 dataset.} to analyze the gold rewritten utterance.
These extracted PA structures are \emph{considered} as gold SRL annotations to measure the accuracy of our conversational SRL parser.
In particular, we evaluate our SRL parser on the micro-averaged F1 over the (\textit{predicate}, \textit{argument}, \textit{label}) tuples.
We find our SRL parser achieves 75.66 precision, 74.47 recall, and 75.06 F$_{1}$.
On the other hand, we use the gold SRL results instead of our SRL parsing results to train and test the model (referred as BERT+Gold-SRL).
From Table~\ref{tab:results}, we can see that all evaluation scores are significantly improved.
This result indicates that the performance of our rewriter model is highly relevant to the SRL parser,
and the performance of our current SRL parser is still far from satisfactory, which we leave for future work.

We also investigate which type of dialogues our model could benefit from incorporating SRL information?
By analyzing the dialogues and our predicted rewritten utterances, we find that the SRL information mainly improves the performance on the dialogues that require information completion.
One omitted information is considered as properly completed if the rewritten utterance recovers the omitted words.
We find the SRL parser naturally offers important guidance into the selection of omitted words.
Examples of rewritten utterances are shown in the Appendix.

Recall that, there is one additional scope option to apply the SRL parser to extract PA structures,
i.e., only working on the last utterance that needs to be rewritten.
We evaluate this option on our dataset (referred as BERT+Partial-SRL) and results are shown in Table~\ref{tab:results}.
We can see that reducing the SRL scope may slightly hurt the performance,
which we think is due to that larger SRL scope could provide additional guidance for the rewriter model.

\section{Conclusions}
In this paper, we introduce a novel SRL-guided framework for enhancing dialogue rewriting. 
For this purpose, we adapted traditional SRL to the conversational scenario by annotating cross-turn annotations on 3,000 dialogues.
Experimental results showed that introducing SRL could significantly improve the rewriting performance without adding extra model parameters.

\bibliographystyle{acl_natbib}

\bibliography{all}

\clearpage
\appendix

\section{Conversational SRL Dataset}
In this section, we first introduce the dialog set that we annotate on and then discuss more details about the annotation.
\subsection{Dialogue Dataset: DuConv}
DuConv is a publicly available knowledge-driven dialogue dataset, focusing on the domain of movies and stars.
It consists of 30k dialogues with 270k dialogue turns and provides a corresponding knowledge graph (KG) .

\subsection{Semantic Roles}
We follow PropBank \cite{carreras2005introduction}, the most widely used standard for annotating predicate-argument structures. It has 32 standard  semantic roles.
By analyzing the conversation dataset, we adopt 9 core semantic roles in our dialogue SRL:\\
$\bullet$ Numbered arguments (\texttt{ARG0}-\texttt{ARG4}): Arguments defining verb-specific roles. Their semantics depends on the verb and the verb usage in a sentence, or verb sense. In general, \texttt{ARG0} stands for the \textit{agent} and \texttt{ARG1} corresponds to the \textit{patient} or \textit{theme} of the proposition, and these two are the most frequent roles. Numbered arguments reflect either the arguments that are required for the valency of a predicate, or if not required, those that occur with high-frequency in actual usage.\\
$\bullet$ Adjuncts: General arguments that any verb may take optionally. In PropBank, there are 13 types of adjuncts, while in our dataset we only consider the most frequent four types of adjuncts, i.e., \texttt{AM-LOC}, \texttt{AM-TMP}, \texttt{AM-PRP} and \texttt{AM-NEG}. Specifically, the locative modifiers (\texttt{AM-LOC}) indicate where the action takes place. The temporal arguments (\texttt{AM-TMP}) show when an action takes place, such as {\small很快} (soon) or {\small马上} (immediately). Note that, the adverbs of frequency (e.g., {\small偶尔} (sometimes), {\small总是} (always)), adverbs of duration (e.g., {\small过两天} (in two days)) and repetition (e.g., {\small又} (again)) are also labeled as \texttt{AM-TMP}. Purpose clauses (\texttt{AM-PRP}) are used to show the motivation for an action. Clauses beginning with {\small为了} (in order to) and {\small因为} (because) are canonical purpose clauses. \texttt{AM-NEG} is used for elements such as `{\small没有}' (not) and `{\small 绝不}' (no longer).

\subsection{Annotation Details} \label{sec:data_anno}
There are two main types of semantic roles: span based \cite{ouchi2018span,tan2018deep} and dependency based \cite{li2019dependency}.
The former involves the start and end boundaries for each component, and the latter only considers the head word in a dependency tree for each component.
We follow the span-based form, which has been adopted by most previous work.

\paragraph{Preprocessing}
For each dialogue session, we first convert it to a paragraph by concatenating each utterance in the dialogue history.
We then use Stanford CoreNLP \cite{manning-EtAl:2014:P14-5} for sentence segmentation, tokenizing, and POS-tagging. We identify verbs by POS tag with heuristics to filter out auxiliary verbs.

\paragraph{Labeling instructions}
We ask five annotators who are familiar with PropBank semantic roles to annotate these dialogue sessions.
Following the span-based annotation standard, annotators label the index ranges for each predicate and its arguments.
In contrast to the standard sentence-level SRL, conversational SRL aims to additionally address the ellipsis and anaphora problems, which frequently occurred in the dialogue scenario.
To this end, the annotators are instructed that a \textit{valid} annotation must satisfy the following criteria: 
(1) the argument should only appear in the current or previous turns; 
(2) the argument should not be assigned to a pronoun unless its reference could not be found in previous turns;
(3) if the argument is the speaker or listener, it should be explicitly assigned to the special token we used to indicate the speaker (i.e., A or B).
(4) in cases when there exit multiple choices for labeling an argument, we select the one that is the closest to the predicate.

\begin{table}[t!] \small
    \centering
    \begin{tabular}{rrr}
    \toprule
        &  Overall Ratio & Cross-turn Ratio  \\
        \midrule
        \texttt{ARG0} & 42.1\% &  22.9\% \\
        \texttt{ARG1} & 40.2\% & 16.9\% \\
        \texttt{ARG2} & 10.1\% & 30.2\% \\
        \texttt{ARG3} & 3.0\% & 24.8\% \\
        \texttt{ARG4} & 0.3\% & 41.4\% \\
        \texttt{AM-TMP} & 3.2\% & 0.3\% \\
        \texttt{AM-LOC} & 1.0\% & 2.1\% \\
        \texttt{AM-PRP} & 0.1\% & 4.0\% \\
    \bottomrule
    \end{tabular}
    \caption{Percent of each type of argument and its cross-turn ratio (shown inside parenthesis).}
    \label{tab:stat}
\end{table}
\paragraph{Statistics}
We annotated 3,000 dialogue sessions from DuConv (33,673 predicates in 27,198 utterances).
Table \ref{tab:stat} analyzes our datasets by listing the percent of each argument type and its cross-turn ratio.
We can see that, for all the three datasets, arguments \texttt{ARG0}, \texttt{ARG1} and \texttt{ARG2} count for the major proportion of the arguments.
For adjunct-type arguments, \texttt{AM-TMP} and \texttt{AM-LOC} appear more than \texttt{AM-PRP}.
It is likely because humans tend to avoid mentioning reasons for simplicity.
Besides, the adjunct-type arguments have very low cross-turn ratios.
This fits our intuition that humans usually mention the time and location when describing an event or a piece of news.

\section{Examples of Model Prediction}
\begin{table}[t!]
\small
\centering
\begin{tabular}{l|l}
& Example \#1 \\
\toprule[0.8pt]
Utterance 1 & $\underline{\text{十一种孤独}}$$_{\textbf{ARG0}}$作者是谁 \\
 & (Who is the author of Eleven Kinds \& of Loneliness ?) \\
Utterance 2 & 理查德耶茨，对吧 \\
& (Richard Yates, right ?) \\
Utterance 3 & 这本书$\underline{\text{讲}}$$_{\textbf{predicate}}$的$\underline{\text{啥}}$$_{\textbf{ARG1}}$ \\
& (What is this book talking about?) \\
\\
\textbf{\textcolor{blue}{Gold:}} & \textbf{\textcolor{blue}{十一种孤独讲的啥}} \\
& (What is Eleven Kinds of Loneliness talking about?)\\
\textbf{\textcolor{blue}{BERT:}} & \textbf{\textcolor{blue}{理查德耶茨讲的啥}} \\
& (What is Richard Yates talking about?)\\
\textbf{\textcolor{blue}{BERT + SRL:}} & \textbf{\textcolor{blue}{十一种孤独讲的啥}} \\
& (What is Eleven Kinds of Loneliness talking about?)\\
\toprule[0.8pt]
& Example \#2 \\
\toprule[0.8pt]
Utterance 1 & $\underline{\text{济南大学}}$$_{\textbf{ARG0}}$\\
& (University of Jinan.) \\
Utterance 2 & 南京一所著名工科强校 \\
& (It is a famous school of engineering in Nanjing.) \\
Utterance 3 & 不, 它$\underline{\text{在}}$$_{\textbf{predicate}}$$\underline{\text{济南}}$$_{\textbf{ARG1}}$ \\
& (No, it is in Jinan.) \\
\\
\textbf{\textcolor{blue}{Gold:}} & \textbf{\textcolor{blue}{不, 济南大学在济南}} \\
& (No, the University of Jinan is in Jinan.)\\
\textbf{\textcolor{blue}{BERT:}} & \textbf{\textcolor{blue}{不, 济南大学在济南大学}} \\
& (No, the University of Jinan is in Jinan University.)\\
\textbf{\textcolor{blue}{BERT + SRL:}} & \textbf{\textcolor{blue}{不，济南大学在济南}} \\
& (No, the University of Jinan is in Jinan.)\\
\toprule[0.8pt]
\end{tabular}
\caption{Examples of multi-turn dialogue. The outputs of our SRL model are annotated in the utterances.}
\label{tab:appendix_examples}
\end{table}
Table~\ref{tab:appendix_examples} gives some running examples of our model predictions.
We can see that with accurate SRL guidence, our model could generate better utterances.

\end{CJK*}
\end{document}